\documentclass[10pt,twocolumn,letterpaper]{article}

\usepackage{iccv}
\usepackage{times}
\usepackage{epsfig}
\usepackage{graphicx}
\usepackage{amsmath}
\usepackage{amssymb}
\usepackage{bbm}
\usepackage{makecell}

\usepackage[pagebackref=true,breaklinks=true,letterpaper=true,colorlinks,bookmarks=false]{hyperref}

\iccvfinalcopy

\begin{document}


\title{Multi-view Action Recognition via \\
 Directed Gromov-Wasserstein Discrepancy}

\author{Hoang-Quan Nguyen, Thanh-Dat Truong, Khoa Luu\\
Computer Vision and Image Understanding Lab \\ 
University of Arkansas, Fayetteville, AR, 72701\\
{\tt\small \{hn016, tt032, khoaluu\}@uark.edu} \\
{\tt\small \url{https://uark-cviu.github.io/}}
}

\maketitle

\begin{abstract}

Action recognition has become one of the popular research topics in computer vision. There are various methods based on Convolutional Networks and self-attention mechanisms as Transformers to solve both spatial and temporal dimensions problems of action recognition tasks that achieve competitive performances. However, these methods lack a guarantee of the correctness of the action subject that the models give attention to, i.e., how to ensure an action recognition model focuses on the proper action subject to make a reasonable action prediction.
In this paper, we propose a multi-view attention consistency method that computes the similarity between two attentions from two different views of the action videos using Directed Gromov-Wasserstein Discrepancy. Furthermore, our approach applies the idea of Neural Radiance Field to implicitly render the features from novel views when training on single-view datasets. Therefore, the contributions in this work are three-fold. 
Firstly, we introduce the multi-view attention consistency to solve the problem of reasonable prediction in action recognition.
Secondly, we define a new metric for multi-view consistent attention using Directed Gromov-Wasserstein Discrepancy. Thirdly, we built an action recognition model based on Video Transformers and Neural Radiance Fields. Compared to the recent action recognition methods, the proposed approach achieves state-of-the-art results on three large-scale datasets, i.e., Jester, Something-Something V2, and Kinetics-400.

\end{abstract}

\section{Introduction}

Automatic action recognition aims to understand human behaviors in a given video and assign either single or multiple action categories to the subjects in that video \cite{sun2020human}. It can be considered one of the fundamental research problems in computer vision with a wide range of applications from camera surveillance, video information retrieval, scene understanding, and human-robot interaction \cite{ranasinghe2016review}. This problem has numerous subcategories, including action classification \cite{kay2017kinetics,goyal2017something,joao2017i3d,limin2017temporal,feichtenhofer2019slowfast} that classifies actions from a sequence of videos, temporal action localization \cite{caba2015activitynet,zhao2017temporal,lin2018bsn,lin2019bmn,zhang2022actionformer} that windows the action segments in a video, 
and spatial-temporal action detection that detects actions via both the spatial and the temporal spaces \cite{gu2018ava,feichtenhofer2019slowfast,sun2018actor,wu2019long}.

\begin{figure}[!t]
\begin{center}
\includegraphics[width=\linewidth]{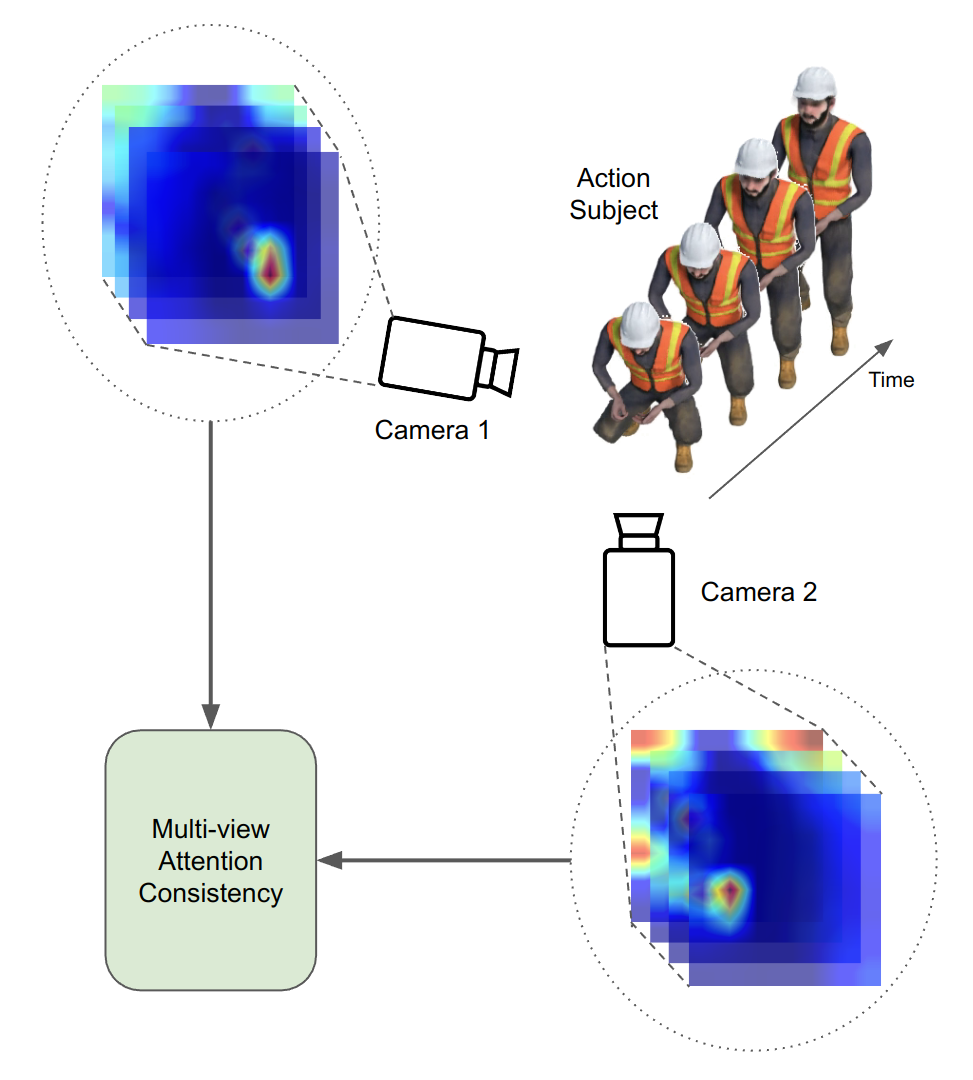}
\end{center}
   \caption{\textbf{The motivation of our method.} Given an action subject with two different camera views, this work aims to ensure that the attention of the model is consistent.}
\label{fig:motivation}
\end{figure}

With the rise of deep neural networks and the growth of large-scale action recognition datasets, many deep learning frameworks have been successfully applied for many action recognition tasks \cite{feichtenhofer2019slowfast, arnab2021vivit}. 
Deploying deep learning models into practical applications., e.g., medicine, security, and finance, etc, the predictions of deep learning models have to be robust and these predictions have to be grounded by evidence or explanation.
However, explaining predictions produced by deep neural networks remains a challenging problem.
To solve this problem, explainable artificial intelligence has been proposed for helping to understand the failures and debug deep neural networks. Commonly, these methods visualize the regions of the model that give attention with respect to the predictions.
In particular,
Class Activation Mapping (CAM) \cite{zhou2016learning} is the early method that visualizes the attention map of the convolutional neural networks using global average pooling layers. 
Later, Grad-CAM \cite{selvaraju2017grad}, produces the attention map by backpropagating the prediction score through the convolutional layers and combining the gradients with the forward features.
Meanwhile, the Transformer \cite{vaswani2017attention} visualizes the explanation using the self-attention weights computed from multi-head self-attention modules.

This work investigates the plausibility of deep neural network models for action recognition. Moreover, the models should automatically correct errors in inference to prevent potentially catastrophic mistakes in future predictions. Our idea to solve this problem is inspired by the cognitive processes of human perception in real-world scenarios.
Specifically, the natural visual perception of humans for recognition tasks is consistent with the spatial transformation of the perception of the subject, particularly the changes in viewing angle. By adopting this approach, we aim to improve the robustness and accuracy of our deep neural network model for action recognition via the consistency of the model from different views. As illustrated in Figure \ref{fig:motivation}, given a sequence of action representation, i.e., a video and two camera views, the model can compute attention volumes from these views and compare their similarity for consistency.

In detail, to guarantee the action recognition model produces a reasonable prediction, the attention of the model to the action subject should be consistent when the view of the video changes. To make the attention of the model consistent with view changes, we first measure the similarity between attention maps from different views of a video. 
Since the changing of view transforms the model's attention, we have to find a metric that compares the structures between the two attentions. Gromov-Wasserstein distance \cite{memoli2011gromov} has succeeded in objective matching and applied to image matching in computer vision \cite{zhang2021deepacg}. However, in image matching, the Gromov-Wasserstein distance computes the topological similarity of images that ignores the motion of the subject, which is crucial in action recognition. We propose a direction-information approach that maintains the motion features from Gromov-Wasserstein distance computing to mitigate this problem.

Furthermore, although there are multi-view action recognition datasets \cite{wang2014cross,shahroudy2016ntu,liu2020ntu} that might be suitable for our approach in the training phase, these datasets are recorded in the laboratory environment. We want to focus on the datasets from real-world contexts even though they consist of single-view videos \cite{materzynska2019jester,goyal2017something,kay2017kinetics}. Thanks to the development of neural fields in visual computing \cite{xie2022neural}, especially Neural Radiance Fields \cite{yen2021inerf}, we can render information from novel views that can obtain the attention of the model from multiple views.

\noindent
\textbf{Contributions of this Work:} Our contributions in this paper are summarized as follows. Firstly, we present the investigation on multi-view attention consistency to solve the problem of reasonable prediction in action recognition. 
Secondly, we propose a new metric for multi-view attention consistency using Directed Gromov-Wasserstein Discrepancy. 
Thirdly, we develop an action recognition model based on Video Transformer and Neural Radiance Fields ideas to obtain attention from different views. 
Finally, 
the experimental results on three large-scale action recognition benchmarks, i.e., Jester \cite{materzynska2019jester}, Something-Something V2 \cite{goyal2017something}, and Kinetics-400 \cite{kay2017kinetics}, have shown the effectiveness of our proposed method.

\section{Related Work}

\begin{figure*}
\begin{center}
\includegraphics[width=\linewidth]{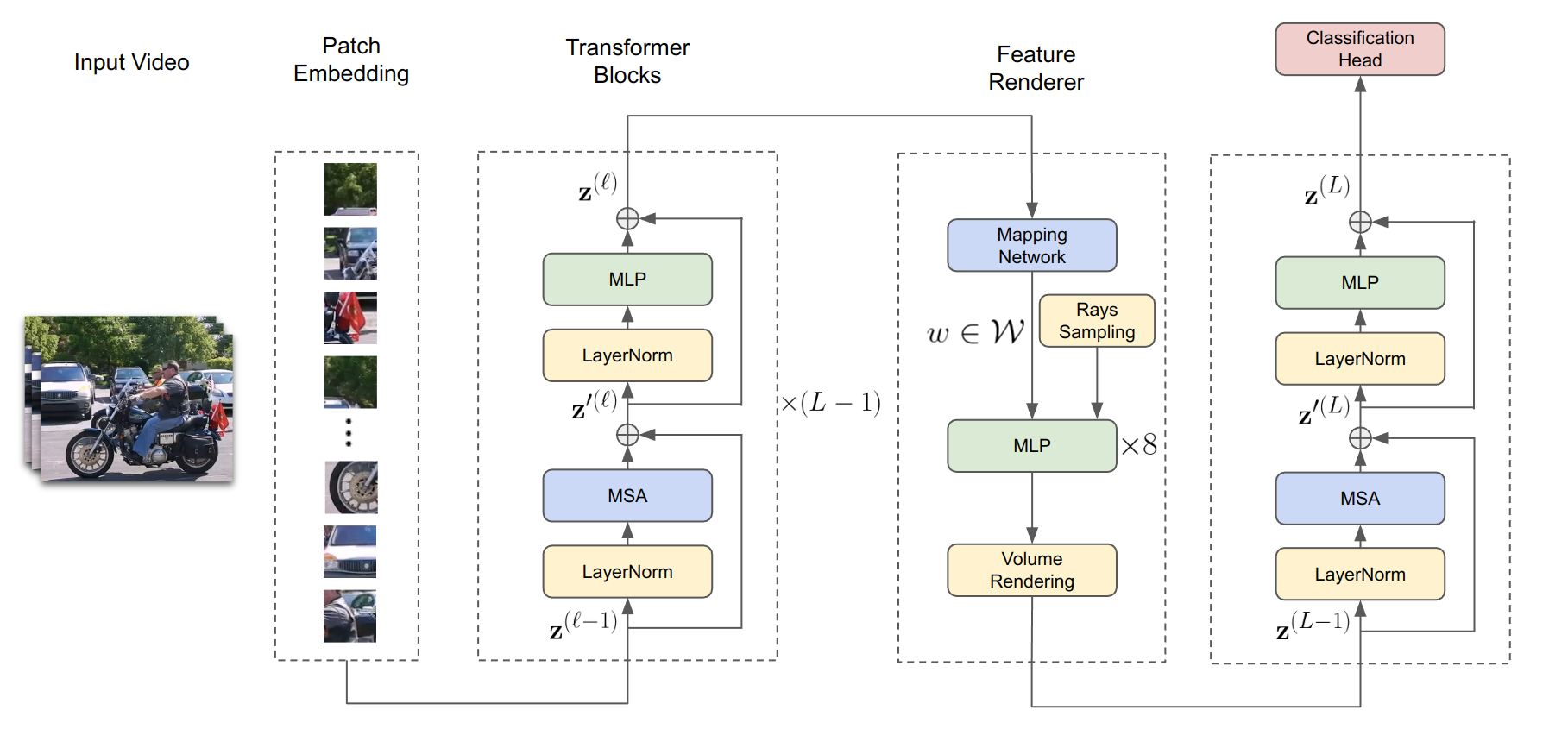}
\end{center}
   \caption{\textbf{The proposed action classification model.} The input video is decomposed into patched and embedded. Then the embedded patches are computed via Transformer blocks. The representation vectors are mapped into the weights of MLP in the feature renderer. With the querying rays, the module renders feature vectors used for the last Transformer block before the classification.}
\label{fig:action_arch}
\end{figure*}

\subsection{Video Action Recognition}

Video understanding is a popular topic in computer vision due to its applications, such as camera surveillance, human behavior analysis, autonomous driving, and robotics. 
Many traditional methods were proposed in the early days using hand-crafted features \cite{wang2011action,wang2013action,peng2014action,lan2015beyond}, particularly Improved Dense Trajectories (IDT) \cite{wang2013action}, which achieved high performance at that time.

With the successful development of deep learning and the existence of large-scale video action datasets such as Kinetics \cite{kay2017kinetics,carreira2018short,carreira2019short}, AVA \cite{gu2018ava}, and Something-Something \cite{goyal2017something}, many deep learning frameworks were introduced. 
Most of the approaches focused on learning spatial-temporal features from the videos. 
Many methods from early progress using 2D CNNs for video problems demonstrated promising results \cite{karpathy2014large,donahue2015long}. 
However, these methods did not outperform the traditional hand-crafted features methods due to their inability to motion handling.

Later methods focused on the motion information, finding an appropriate way to describe the temporal relationship between frames to improve the performance of CNN-based video action recognition. 
There are two categories of approaches. 
The first group of methods utilized the optical flow \cite{horn1981determining} of the video to describe scene movements.
\cite{simonyan2014two} proposed two-stream networks, including spatial and temporal streams. 
The spatial stream extracted features from raw video frames to capture visual appearance information. 
The temporal stream extracted features from optical flow images to capture the motion content of the video.

The second group of methods for video action recognition includes 3D-CNN-based methods. 
Inspired by \cite{ji20123d}, C3D \cite{tran2015learning} applied 3D convolutions to model the spatial and temporal features together. 
However, 3D networks are hard to optimize due to their large number of parameters.

The rise of Transformer approaches plays an important role due to their competitive accuracy and maintained computational resources compared to convolutional methods. 
The early video transformer model ViViT \cite{arnab2021vivit} factorized the spatial and temporal dimensions of the video to handle spatial-temporal information from a long sequence of frames. 
TimeSformer \cite{bertasius2021space} demonstrated that separating spatial and temporal attention within each block increases the accuracy and performance of the standard video Transformer. 
Video Swin Transformer \cite{liu2022video} computed self-attention locally via window shifting for better computational speed and accuracy.
DirecFormer \cite{truong2022direcformer} introduced a directed self-attention mechanism between frames for motion robustness.
The spatial-temporal transformer framework was also applied in video group activity recognition \cite{chappa2023spartan,chappa2023sogar,chappa2023react}.
Moreover, a geometric-based Transformer was proposed for cross-view action recognition \cite{truong2023cross}.
Meanwhile, a dynamic graph-based Transformer was applied to analyze the spatial-temporal relationship between subjects from multiple camera views \cite{quach2021dyglip}.

\subsection{Attention Map Consistency}

There are many approaches considering the consistency inference of the vision model. 
Some methods focused on the deep equivariance indicating that the representation of the input should follow the transformation of that input \cite{cohen2016group,dieleman2016exploiting,worrall2017harmonic,marcos2017rotation,worrall2018cubenet}. 
Another group of approaches collaborated on different neural network modules that transfer the learned information between these networks \cite{zagoruyko2016paying,tarvainen2017mean,niu2019multi,qiao2018deep}.

The consistency of visual attention maps has recently received interest in computer vision. 
There have been several works on evaluating and optimizing models from the consistency between multiple attention maps for visual tasks.
Some works focused on the consistency between attention maps under different augmentation and masking procedures \cite{guo2019visual,li2020unsupervised,cugu2022attention,prabhu2022adapting}. 
At the same time, other methods inspected the consistency between attention maps from different layers \cite{wang2019sharpen,xu2022exploiting} or different networks \cite{zagoruyko2016paying}.
FALCON \cite{truong2023falcon} introduced contrastive learning between attention maps to model the background shift problem in continual semantic segmentation.
Meanwhile, ATCON \cite{mirzazadeh2023atcon} utilized the consistency between multiple attention map methods.
Unlike those methods, we focus on the attention consistency between two views of an action subject.

\subsection{Neural Radiance Fields}

The objective of the Neural Radiance Fields (NeRF) is to synthesize views of a scene by querying the position and direction of points along camera rays and using classic volume rendering techniques to project the output colors and densities into an image \cite{mildenhall2020nerf}. There is a rise in NeRF variants for different tasks. 
NeRF++ \cite{zhang2020nerf++} improves the quality of unbounded 3D scenes using an inverted-sphere background parameterization. 
Some frameworks propose back-propagating into camera parameters to enable camera pose estimation \cite{wang2021nerfmm,lin2021barf,yen2021inerf}.
GRF \cite{trevithick2021grf}, and PixelNeRF \cite{yu2021pixelnerf} synthesize novel images from prior information by extracting features from images of a scene.
D-NeRF \cite{pumarola2021d}, NR-NeRF \cite{tretschk2021non}, and Nerfies \cite{park2021nerfies} perform non-rigidly deformation reconstruction for dynamic scenes.
NeRF framework is also applied to high-resolution image generation by rendering low-resolution feature maps and progressively applying upsampling in 2D \cite{gu2022stylenerf}.
In this work, we adopt the idea of NeRF as a layer for action recognition to model the low-resolution features from multiple camera views.

\section{Our Proposed Method}

Let $x \in \mathbb{R}^{T \times H \times W \times 3}$ be an input video where $T$, $H$, and $W$ are the number of frames, height, and width of a video, $\beta \in \mathbb{R}$ be a moving angle of the video camera, $F: \mathbb{R}^{T \times H \times W \times 3} \times \mathbb{R} \rightarrow \mathbb{R}^{C} \times \mathbb{R}^{t \times h \times w}$ where $C$ is the number of action classes be an action classification function mapping a video $x$ with camera angle $\beta$ to an action prediction $y \in \mathbb{R}^{C}$ and a spatial-temporal attention volume $a \in \mathbb{R}^{t \times h \times w}$ corresponding to the video where ($t$, $h$, $w$) is the shape (time, height, width) of the attention. Our goal is to learn a deep neural network to classify the actions and define a metric that computes the consistency between the attention volumes of the model from different camera views.

In this section, we will present the fundamentals of the Video Transformer \cite{arnab2021vivit,bertasius2021space,liu2022video} in Section \ref{sec:video_transformer}. 
Then in Section \ref{sec:nerf_attn}, we will talk about the usage of Neural Radiance Fields \cite{mildenhall2020nerf} for attention volumes in different camera views.
Finally, Section \ref{sec:attn_consistent_sim} proposes the attention-consistent similarity computing using directed Gromov-Wasserstein Discrepancy \cite{peyre2016gromov}.

\subsection{Video Transformer}
\label{sec:video_transformer}

In this work, we follow the Video Transformer frameworks for action recognition. In detail, the video $x$ is decomposed into $P$ patches and embedded via positional encoding. Then the embedded patches $\{\bold{z}^{(0)}_i\}_{i=1}^{P}$ are feed-forwarded into a Transformer encoder \cite{vaswani2017attention}.

The Transformer encoder \cite{vaswani2017attention} consists of $L$ blocks of alternating layers of multi-head self-attention (MSA) and MLP.
The layer normalization (LN) \cite{ba2016layer} is applied before each block and the residual connection is applied after each layer.
\begin{align}
    \bold{z}^{\prime(\ell)}_p 
    &= \text{MSA}(\text{LN}(\bold{z}^{(\ell-1)}_p)) + \bold{z}^{(\ell-1)}_p , 
    && \ell = 1 \ldots L \\
    \bold{z}^{(\ell)}_p 
    &= \text{MLP}(\text{LN}(\bold{z}^{\prime(\ell)}_p)) + \bold{z}^{\prime(\ell)}_p , 
    && \ell = 1 \ldots L \label{eq:mlp_res}
\end{align}

At each multi-head self-attention layer, a query, key, and value vector is computed for each patch from the representation $\bold{z}^{(\ell-1)}_p$ encoded from the previous block.
\begin{align}
    \bold{q}^{(\ell,a)}_p = W^{(\ell,a)}_{Q} \text{LN}(\bold{z}^{(\ell-1)}_p) \\
    \bold{k}^{(\ell,a)}_p = W^{(\ell,a)}_{K} \text{LN}(\bold{z}^{(\ell-1)}_p) \\
    \bold{v}^{(\ell,a)}_p = W^{(\ell,a)}_{V} \text{LN}(\bold{z}^{(\ell-1)}_p)
\end{align}
Then the self-attention weights vector $\pmb{\alpha}^{(\ell,a)}_{p} \in \mathbb{R}^{P}$ is computed from query vector $\bold{q}^{(\ell,a)}_{p}$ and the concatenation of key vectors $\{\bold{k}^{(\ell,a)}_{p^{\prime}}\}^{P}_{p^{\prime}=1}$ as:
\begin{align}
    k^{(\ell,a)} &= \text{concat}([\bold{k}^{(\ell,a)}_{1}, \bold{k}^{(\ell,a)}_{2}, \dots , \bold{k}^{(\ell,a)}_{P}]) \\
    \pmb{\alpha}^{(\ell,a)}_{p} &= \text{softmax}\left(
    \frac{\bold{q}^{(\ell,a)}_{p}}{\sqrt{d}}^{\top} 
    \cdot k^{(\ell,a)}
    \right)
\end{align}
where $d$ is the dimension of the query and key vectors. The encoding $\bold{z}^{\prime(\ell)}_p$ is computed by firstly calculating the weighted sum of the value vectors for each self-attention head:
\begin{equation}
    \bold{s}^{(\ell,a)}_p = \sum_{i=1}^{P}{\alpha^{(\ell,a)}_{p,i} \bold{v}^{(\ell,a)}_i}
\end{equation}
Then these vectors all over the heads are concatenated and projected with residual connection:
\begin{align}
    \bold{s}^{(\ell)}_p &= \text{concat}\left(\left[\begin{array}{c}
         \bold{s}^{(\ell,1)}_p \\
         \bold{s}^{(\ell,2)}_p \\
         \vdots \\
         \bold{s}^{(\ell,A)}_p
    \end{array}\right] \right)\\
    \bold{z}^{\prime(\ell)}_p &= W^{(\ell)}_{O} \bold{s}^{(\ell)}_p + \bold{z}^{(\ell-1)}_p
\end{align}
Hence, $\bold{z}^{\prime(\ell)}_p$ is passed through an MLP with residual connection as Equation \ref{eq:mlp_res} to obtain encoding $\bold{z}^{(\ell)}_p$.

\subsection{Neural Radiance Field for Attention Volume}
\label{sec:nerf_attn}

To obtain the attention map from a novel camera view, a trivial approach is explicitly rendering a video from that angle using NeRF frameworks \cite{mildenhall2020nerf}. However, this approach requires a large computational resource and is not applicable to our action recognition method. To reduce the computational cost, we implicitly change the attention map without modifying the video, 
i.e. render a feature map that has a lower resolution based on the StyleNeRF idea \cite{gu2022stylenerf}.

In detail, given an input video $x \in \mathbb{R}^{T \times H \times W \times 3}$, we compute the feature volume of $x$ using Transformer $z = f(x)$, $z \in \mathbb{R}^{t \times h \times w \times d}$, as discussed in Section \ref{sec:video_transformer}. 
Following the idea of StyleGAN \cite{karras2020analyzing}, we map the feature volume $z$ to style vectors $w \in \mathcal{W}$ and modulate to the weight matrix of the MLP layers in the NeRF module. 
Next, given an camera parameter matrix $c \in \mathbb{R}^{3 \times 4}$, we render a low-resolution feature volume $z^{g} = g(z, c)$, $z^{g} \in \mathbb{R}^{t \times h \times w \times d}$. 
A feature vector $\bold{z}^{g}_{\bold{r}} \in \mathbb{R}^{d}$ is rendered from a ray $\bold{r}$ using volumetric rendering calculating the cumulation weighted feature vectors obtained from MLPs. In theory, the feature $\bold{z}^{g}_{\bold{r}}$ can be compute as:
\begin{equation}
    \label{eq:integral_volume_rendering}
    \bold{z}^{g}_{\bold{r}} = \int_{0}^{\infty} T(u) \sigma(\bold{r}(u)) \bold{z}^{g}_{\bold{r}}(u) du
\end{equation}
where $T(u) = \exp \left(-\int_{0}^{u} \sigma(\bold{r}(v)) dv \right)$ is the probability that the ray travels from the origin of the camera to $u$ without hitting any other particle, $\sigma(\bold{r}(u))$ is the density of point $u$ on the ray $\bold{r}$ \cite{mildenhall2020nerf}.
Then, we can discretize the Equation \ref{eq:integral_volume_rendering} using a discrete set of samples on ray $\bold{r}$ and apply the discrete volumetric rendering based on \cite{max1995optical}:
\begin{equation}
    \bold{z}^{g}_{\bold{r}} = \sum_{i=1}^{N} T_{i} (1 - \exp \left( \sigma_{i} \delta_{i} \right)) \bold{z}^{g}_{\bold{r}, i}
\end{equation}
where $T_{i} = \exp \left(- \sum_{j=1}^{i-1} \sigma_{j} \delta_{j} \right)$ and $\delta_{i}$ is the distance between two adjacent points on the ray.

Finally, we compute the feature for action classification and visual attention using the low-resolution feature volume $z^{g}$ via another Transformer block. The model can be visualized as Figure \ref{fig:action_arch}. Here we use Video Swin Transformer \cite{liu2022video} as a video extractor. Note that this framework can be applied to other Transformer architectures such as ViViT \cite{arnab2021vivit} and TimeSformer \cite{bertasius2021space}. The difference of Video Swin Transformer \cite{liu2022video} from the others \cite{arnab2021vivit,bertasius2021space} is the multi-head self-attention computing between patches in the shifted windows instead of computing the self-attention between the patches in the whole video.

\subsection{Attention Consistent Similarity}
\label{sec:attn_consistent_sim}

Observe that from the same action, when we capture it from different views, the attention maps might differ. However, due to the view consistency, the model focuses on the same points, thus the motion and the structure of the attention maps should be the same. Hence, to compare the similarity between two attention maps, we have to define a function that is robust to the transition of the camera which causes translation to attention maps. We find that Gromov-Wasserstein (GW) distance \cite{memoli2011gromov} has this property.

Recall the Gromov-Wasserstein distance \cite{memoli2011gromov} compares distributions by computing the similarity between the metrics defined within each of the spaces, meaning the structures of the distributions. 
Given $n$ samples of the compared distributions $p$ and $\bar{p}$, we can discrete the formulation of the Gromov-Wasserstein distance into a discrepancy function \cite{peyre2016gromov} using a distance matrix $D \in \mathbb{R}^{n \times n}$ between samples and a probability vector $\bold{p} \in \mathbb{R}^n, \sum_{i}{\bold{p}_i} = 1$ for each space. 
Then the Gromov-Wasserstein discrepancy is formulated as:
\begin{equation}
\begin{aligned}
\label{eq:gw_discrepacy}
    \text{GW}(D, \bar{D}, \bold{p}, \bold{\bar{p}}) 
    &= \min_{T \in \mathcal{U}_{\bold{p}, \bold{\bar{p}}}} \mathcal{E}_{D, \bar{D}}(T) \\
    &= \min_{T \in \mathcal{U}_{\bold{p}, \bold{\bar{p}}}} \sum_{i,j,k,l}{\mathcal{L}(D_{i,k}, \bar{D}_{j,l}) T_{i,j} T_{k,l}}
\end{aligned}
\end{equation}
where $\mathcal{U}_{\bold{p}, \bold{\bar{p}}} = \{T \in (\mathbb{R_{+}})^{n \times n}; T \mathbbm{1}_{n} = \bold{p}, T^{\top} \mathbbm{1}_{n} = \bold{\bar{p}} \}$ is a set of all coupling matrices $T$ between $\bold{p}$ and $\bold{\bar{p}}$, $D$ and $\bar{D}$ are intra-distance matrices of the two distributions, and $\mathcal{L}(u, v)$ is a loss function between the two scalars.

To apply the Equation \ref{eq:gw_discrepacy} for consistency comparison of two attention $a_1, a_2 \in \mathbb{R}^{t \times h \times w}$, we compute intra-distance matrices $D_1, D_2 \in \mathbb{R}^{thw \times thw}$ between points in each attention volume. After reshaping the attentions into probability vectors $\bold{p_1}, \bold{p_2} \in \mathbb{R}^{thw}$, we have an attention consistency loss function as Equation \ref{eq:gw_loss}.
\begin{equation}
\label{eq:gw_loss}
    \mathcal{L}_{\text{GW}}(a_1, a_2) = \text{GW}(D_1, D_2, \bold{p_1}, \bold{p_2})
\end{equation}

\noindent
\textbf{Directed Gromov-Wasserstein Discrepancy} 
This application of Gromov-Wasserstein distance is similar to \cite{zhang2021deepacg,solomon2016entropic} for objective mapping. 
In regular, the intra-distance matrix is calculated using Euclidean distance and the loss function $\mathcal{L}(u, v)$ is the squared error loss $\mathcal{L}_2(u, v) = \frac{1}{2} (u - v)^2$. 
However, in our case, the points in the attention have position information that represents the motion and structural information of the video. 
If we directly apply the Euclidean distance, the information such as motions and spatial structure of the attention will be omitted. 
To alleviate this problem, we add the direction information to the intra-distance matrix. 
In particular, we replace the intra-distance matrix $D$ with an intra-vector matrix $V \in \mathbb{R}^{thw \times thw \times 3}$ that calculates the vectors between two points in the attention volume. 
Then the $\mathcal{L}(u, v)$ from Equation \ref{eq:gw_discrepacy} is defined as a cosine similarity function $\mathcal{L}_{\text{cosine}}(\bold{u}, \bold{v})$ scaled to be in the range $[0, 1]$ where $0$ means the two vectors are in the same direction and $1$ means the two vectors are in the opposite direction. Hence the directed Gromov-Wasserstein (DGW) discrepancy is formulated as:
\begin{equation}
\begin{aligned}
\label{eq:dgw_discrepacy}
    \text{DGW}(V, \bar{V}, \bold{p}, \bold{\bar{p}}) 
    &= \min_{T \in \mathcal{U}_{\bold{p}, \bold{\bar{p}}}} \sum_{i,j,k,l}{\mathcal{L}_{\text{cosine}}(V_{i,k}, \bar{V}_{j,l}) T_{i,j} T_{k,l}}
\end{aligned}
\end{equation}
where $\mathcal{L}_{\text{cosine}}(\bold{u}, \bold{v})$ is defined as:
\begin{equation}
    \mathcal{L}_{\text{cosine}}(\bold{u}, \bold{v}) 
    = \frac{1}{2} \left( 1 - \frac{\bold{u}^{\top}\bold{v}}{\|\bold{u}\|\|\bold{v}\|} \right)
\end{equation}
Hence, the attention consistency loss function is now computed as Equation \ref{eq:dgw_loss}.
\begin{equation}
\label{eq:dgw_loss}
    \mathcal{L}_{\text{DGW}}(a_1, a_2) = \text{DGW}(V_1, V_2, \bold{p_1}, \bold{p_2})
\end{equation}

\begin{figure*}
\begin{center}
\includegraphics[width=\linewidth]{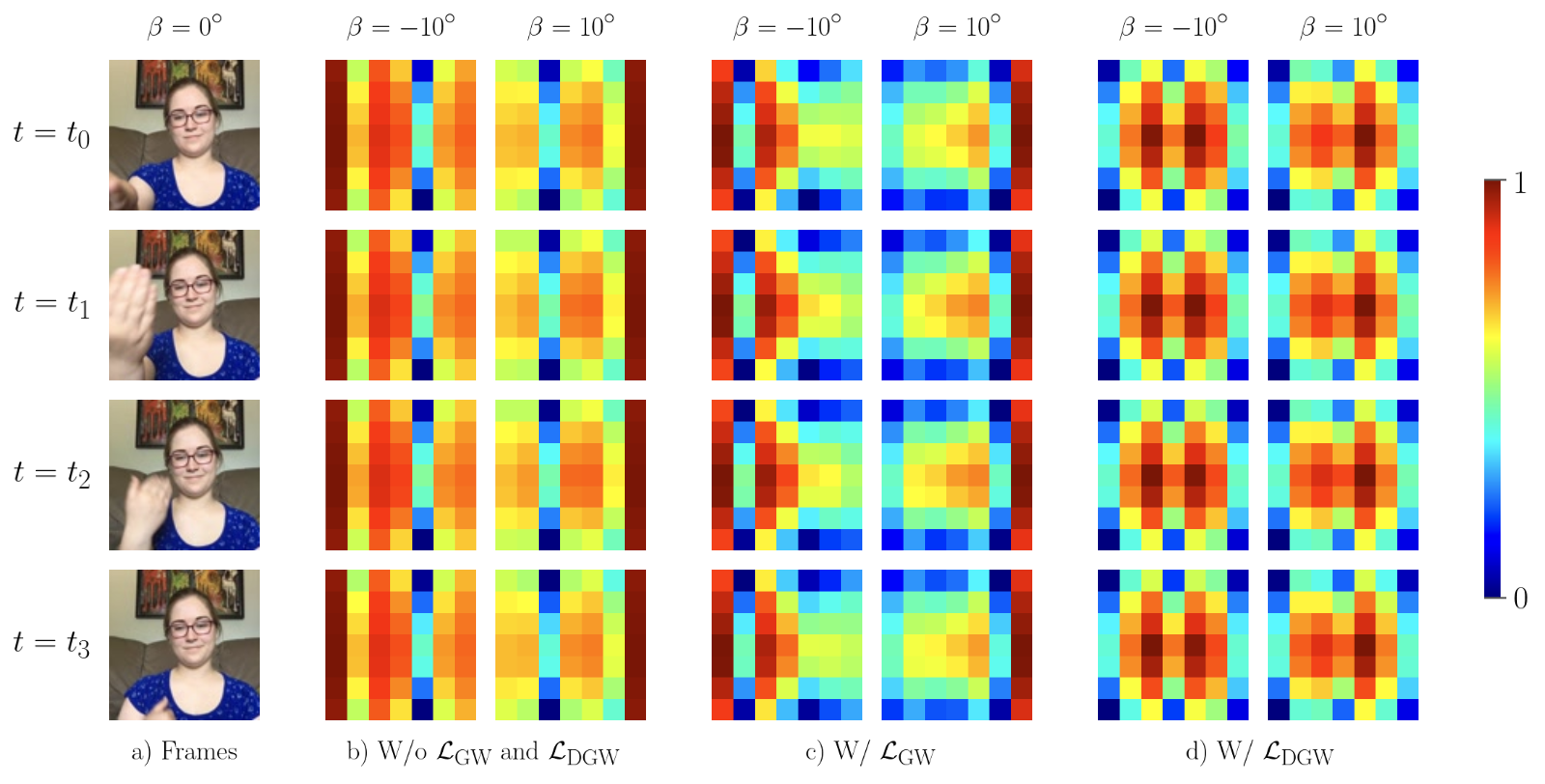}
\end{center}
   \caption{The attention visualization of four frames (a) from two camera angles $\beta=-10^{\circ}$ and $\beta=10^{\circ}$ on the three settings: (b) without $\mathcal{L}_{\text{GW}}$ and $\mathcal{L}_{\text{DGW}}$, (c) with $\mathcal{L}_{\text{GW}}$, and (d) with $\mathcal{L}_{\text{DGW}}$.}
\label{fig:attn_vis}
\end{figure*}

\noindent
\textbf{Gromov-Wasserstein Discrepancy Optimization}
Computing the Gromov-Wasserstein discrepancy is a quadratic programming problem that is difficult to solve. Even those, we can approximately solve this problem using entropic regularization as Equation \ref{eq:entropic_gw} for more efficient optimization \cite{peyre2016gromov}.
\begin{equation}
\label{eq:entropic_gw}
    \text{GW}_{\epsilon}(D, \bar{D}, \bold{p}, \bold{\bar{p}}) 
    = \min_{T \in \mathcal{U}_{\bold{p}, \bold{\bar{p}}}} \mathcal{E}_{D, \bar{D}}(T) 
    - \epsilon E(T) \\
\end{equation}
\begin{equation}
    E(T) = - \sum_{i,j} T_{i,j} \log(T_{i,j})
\end{equation}
Then, it can be solved via projected gradient descent methods. This can be treated as the Optimal Transport problem and can be solved by the Sinkhorn-Knopp algorithm \cite{cuturi2013sinkhorn}. The directed Gromov-Wasserstein discrepancy can be solved similarly to Equation \ref{eq:entropic_gw}.

Finally, the total loss function of our method is defined as in Equation \ref{eqn:total_final_loss}.
\begin{equation} \label{eqn:total_final_loss}
    \mathcal{L}_{\text{total}} = \lambda_{\text{cls}} \mathcal{L}_{\text{cls}} + 
    \lambda_{\text{DGW}} \mathcal{L}_{\text{DGW}}
\end{equation}
where $\mathcal{L}_{\text{cls}}$ is the cross-entropy loss of the classification; 
$\lambda_{\text{cls}}$ and $\lambda_{\text{DGW}}$ are the hyperparameters controlling the importance between losses.

\section{Experiments}

\begin{figure}
\begin{center}
\includegraphics[width=\linewidth]{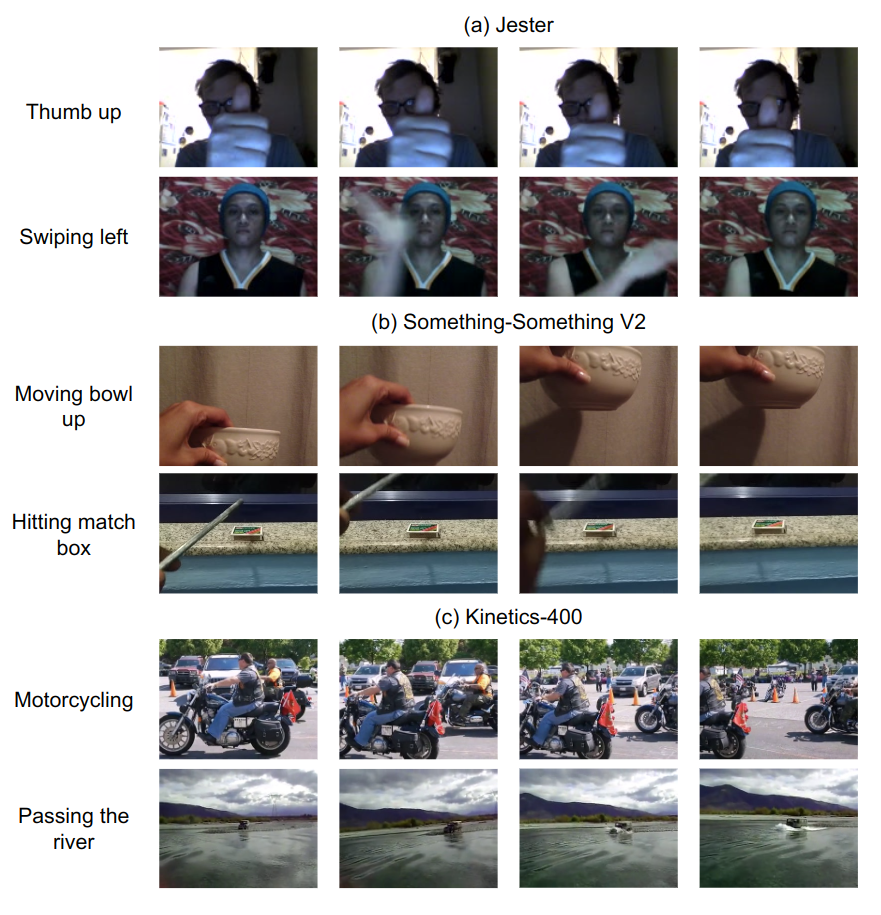}
\end{center}
   \caption{\textbf{The samples of three large-scale action recognition datasets:} (a) Jester \cite{materzynska2019jester}, (b) Something-Something V2 \cite{goyal2017something}, and (c) Kinetics-400 \cite{kay2017kinetics}.}
\label{fig:datasets}
\end{figure}

In this section, we present our experiments on three popular action recognition datasets, including Jester \cite{materzynska2019jester}, Something-Something V2 \cite{goyal2017something}, and Kinetics-400 \cite{kay2017kinetics}. 
Firstly, we describe our implementation details and the datasets used for our experiments. The datasets samples are illustrated as Figure \ref{fig:datasets}. For all the comparisons of the methods, we evaluate the performances by Top-1 and Top-5 recognition accuracy. 
Secondly, we analyze our quantitative results with different settings, as shown in the ablation study on the Jester dataset. We also visualize the quality of our model via different consistency losses to illustrate the robustness of the model to different camera angles.
Lastly, we compare our evaluation results on the Kinetics-400 \cite{kay2017kinetics} and Something-Something V2 \cite{goyal2017something} datasets to prior state-of-the-art methods.

\subsection{Datasets}

\noindent
\textbf{Jester} \cite{materzynska2019jester} 
The dataset is a large-scale video gesture dataset. It consists of 148,092 videos of humans performing 27 types of basic, pre-defined hand gestures in front of a laptop camera or webcam. The dataset contains 118,562 videos for training, 14,787 videos for validation, and 14,743 videos for testing. 
In our experiment, we follow the evaluation protocol of the previous papers to report the accuracy of the model on the validation set.

\noindent
\textbf{Something-Something V2} \cite{goyal2017something} 
It is a dataset of humans performing actions with everyday objects. The dataset includes 220,847 videos from 174 classes, with 168,913 videos for training, 24,777 videos for validation, and 27,157 videos for testing.
Similar to the previous works, we report the accuracy of the model on the validation set.
Both Jester and Something-Something V2 datasets are under licenses registered by Qualcomm Technologies Inc. that are publicly available for academic research.

\noindent
\textbf{Kinetics-400} \cite{kay2017kinetics} 
The dataset contains 400 human action classes, with at least 400 videos downloaded from YouTube, and each video lasts for around 10 seconds. In detail, the dataset consists of 306,245 videos, including 234,619 videos for training and 19,761 videos for validation. 
The dataset covers different types of actions: Person Actions (e.g. drinking, smoking, singing, etc.), Person-Person Actions (e.g. shaking hands, kissing, wrestling, etc.), and Person-Object Actions (e.g., washing dishes, opening a bottle, making a sandwich, etc.).
The Kinetics dataset is licensed by Google Inc. under a Creative Commons Attribution 4.0 International License.

\subsection{Implementation Details}

The model is built based on Video Swin Transformer \cite{liu2022video} and consists of 4 stages of Swin Transformer. The number of Swin Transformer blocks for each stage is $(2, 2, 18, 2)$. The feature computed after the third stage is used for rendering a low-resolution feature via the NeRF module. Then, this feature is computed via the last Swin Transformer stage for classification and visual attention. The low-resolution feature has a size of $4 \times 7 \times 7 \times 1024$. The NeRF module has $8$ MLP layers with Leaky ReLU activations \cite{maas2013rectifier} and the width of each layer is $1024$.

The attention for consistency computing is obtained from the last Transformer block. Initially, the attention has the size of $A \times twh \times twh$ where $A = 32$ is the number of self-attention heads. We reshape the attention into $A \times twh \times t \times w \times h$ and average the attention along the second dimension to derive $A$ attention volumes each sizes $t \times w \times h$. Then we compute the average attention consistency loss $\mathcal{L}_{\text{GW}}$ or $\mathcal{L}_{\text{DGW}}$ of $A$ pairs of attention volumes between two camera views.

The input video consists of $T = 8$ frames and the resolution of each frame is $224 \times 224$ ($H = W = 224$). We train the model with a batch size of $16$. We employ the AdamW \cite{loshchilov2017decoupled} optimizer and train for $30$ epochs with the learning rate $\text{lr}=10^{-4}$. 
We set the control hyperparameters of losses to $1.0$, i.e. $\lambda_{\text{cls}} = \lambda_{\text{DGW}} = 1.0$.
In novel views rendering training, we sample the camera angle in the range $[-10^{\circ}, 10^{\circ}]$ turning horizontally around an imagined center of the scenes in videos.

For inference, similar to the Video Swin Transformer \cite{liu2022video} which follows \cite{arnab2021vivit}, our method uses $4 \times 3$ views ensembling, where a video is uniformly sampled as four clips in the temporal dimension, and the shorter spatial side of each frame is scaled to $224$ pixels. We take the three crops of size $224 \times 224$ that cover the longer side. Then we obtain the final result by averaging the scores of all the views.

\subsection{Ablation Study}

\begin{table}
\setlength{\tabcolsep}{2.5pt}
\begin{center}
\begin{tabular}{l|c|c|c|cc}
\hline
\textbf{Model} & \textbf{\thead{Novel\\Views}} & $\mathcal{L}_{\text{GW}}$ & $\mathcal{L}_{\text{DGW}}$ & \textbf{Top-1} & \textbf{Top-5} \\
\hline
I3D \cite{carreira2017quo} & - & - & - & 91.46 & 98.67 \\
3D-SqueezeNet \cite{iandola2016squeezenet} & - & - & - & 90.77 & - \\
ResNet-50 \cite{he2016deep} & - & - & - & 93.70 & - \\
ResNet-101 \cite{he2016deep} & - & - & - & 94.10 & - \\
3D-MobileNetV2 \cite{kopuklu2019resource} & - & - & - & 94.59 & - \\
ResNeXt-101 \cite{xie2017aggregated} & - & - & - & 94.89 & - \\
ViVit \cite{arnab2021vivit} & - & - & - & 81.70 & 93.80 \\
TimeSformer \cite{bertasius2021space} & - & - & - & 94.14 & 99.19 \\
Swin-B \cite{liu2022video} & - & - & - & 96.56 & 99.82 \\
\hline\hline
Ours &  &  &  & 96.50 & 99.83 \\
Ours & \checkmark &  &  & 96.69 & 99.80 \\
Ours & \checkmark & \checkmark &  & 96.77 & 99.84 \\
\textbf{Ours} & \checkmark &  & \checkmark & \textbf{96.94} & \textbf{99.87}\\
\hline
\end{tabular}
\end{center}
\caption{\textbf{Ablation Study on Jester dataset \cite{materzynska2019jester}.} We evaluate the performance of attention consistency in three settings: without attention consistency loss, with GW loss, and with directed GW loss.}
\label{table:jester}
\end{table}

To demonstrate the effectiveness of our directed Gromov-Wasserstein loss $\mathcal{L}_{\text{DGW}}$, we evaluate the performance of the action recognition model in the three settings, i.e., training without attention consistency loss, with regular Gromov-Wasserstein loss $\mathcal{L}_{\text{GW}}$, and with directed Gromov-Wasserstein loss $\mathcal{L}_{\text{DGW}}$. We also evaluate the Video Swin Transformer \cite{liu2022video} and our model training without novel view rendering. The results are shown in Table \ref{table:jester}.

\noindent
\textbf{Effectiveness of Novel View Rendering Training} 
We compare the evaluations of our model when we train with a single view for each video, i.e., the camera angle is at $\beta=0^{\circ}$, and with the multiple views for each video, i.e., sampling multiple camera angles to render novel views implicitly. As in Table \ref{table:jester}, the novel-views training strategy increases the accuracy from $96.50\%$ to $96.69\%$.

\noindent
\textbf{Effectiveness of Directed Gromov-Wasserstein Discrepancy}
Illustrated as Table \ref{table:jester}, with the Gromov-Wasserstein loss $\mathcal{L}_{\text{GW}}$, the performance of our model has been improved. Furthermore, the directed Gromov-Wasserstein loss $\mathcal{L}_{\text{DGW}}$ gets better result compared to the original Gromov-Wasserstein loss $\mathcal{L}_{\text{GW}}$ meaning the directed Gromov-Wasserstein loss $\mathcal{L}_{\text{DGW}}$ has maintained the motion information and the spatial structure of the attention the model focuses on when computing attention consistency.

\subsection{Multi-view Attention Visualization Analysis}

To demonstrate the robustness of attention consistency losses, we use a video on the validation set of Jester to visualize the attention of two different camera views from a video of the action "Pulling Hand In", as shown in Figure \ref{fig:attn_vis}, in the three settings, i.e., without the consistency losses $\mathcal{L}_{\text{GW}}$ and $\mathcal{L}_{\text{DGW}}$, with Gromov-Wasserstein loss $\mathcal{L}_{\text{GW}}$, and with directed Gromov-Wasserstein loss $\mathcal{L}_{\text{DGW}}$.
We analyze how similar the two attentions are from camera angle $\beta=-10^{\circ}$ and $\beta=10^{\circ}$ turning around the scene's center within four frames. As illustrated in Figure \ref{fig:attn_vis}, with the directed Gromov-Wasserstein loss $\mathcal{L}_{\text{DGW}}$, the model gives attention to the action subject better than the others.

\subsection{Comparison to State-of-the-art Methods}

\begin{table}
\begin{center}
\begin{tabular}{l|cc}
\hline
\textbf{Model} & \textbf{Top-1} & \textbf{Top-5} \\
\hline
I3D \cite{carreira2017quo} & 74.20 & 91.30 \\
SlowFast R101+NL \cite{feichtenhofer2019slowfast} & 79.80 & 93.90 \\
X3D-XXL \cite{feichtenhofer2020x3d} & 80.00 & 94.50 \\
TimeSformer-L \cite{bertasius2021space} & 80.70 & 94.70 \\
MViT-B \cite{fan2021multiscale} & 81.20 & 95.10 \\
ViViT-L \cite{arnab2021vivit} & 81.70 & 93.80 \\
Swin-B ImageNet-1K \cite{liu2022video} & 80.60 & 94.60 \\
Swin-B ImageNet-21K \cite{liu2022video} & 82.70 & 95.50 \\
\hline\hline
\textbf{Ours} & \textbf{83.49} & \textbf{96.17} \\
\hline
\end{tabular}
\end{center}
\caption{\textbf{Comparison to state-of-the-art methods on Kinetics-400 dataset \cite{kay2017kinetics}.} The proposed evaluation result is trained using directed Gromov-Wasserstein loss $\mathcal{L}_{DGW}$.}
\label{table:k400}
\end{table}

\noindent
\textbf{Kinetics-400} 
Table \ref{table:k400} presents the performance of our proposed approach evaluated on Kinetics-400 compared to prior state-of-the-art approaches.
In this experiment, our model uses the pretrained model on ImageNet-21K \cite{deng2009imagenet} similar to the training protocol of Video Swin Transformer \cite{liu2022video}. We use the directed Gromov-Wasserstein loss $\mathcal{L}_{\text{DGW}}$ for training. As in Table \ref{table:k400}, our result outperforms other candidates with the Top-1 accuracy sitting at $83.49\%$ and the Top-5 accuracy at $96.17\%$.

\noindent
\textbf{Something-Something V2} 
Table \ref{table:ssv2} illustrates the comparisons to the state-of-the-art methods, including Convolution and Transformer approaches, on Something-Something V2. Similar to \cite{liu2022video}, the pretrained model on Kinetics-400 \cite{kay2017kinetics} is used in this experiment. As shown in Table \ref{table:ssv2}, our method achieves state-of-the-art performance compared to the prior approaches. The Top-1 accuracy of our method is $70.74\%$, $1.14\%$ higher than Swin-B \cite{liu2022video}, our base model.
The effectiveness of the proposed multi-view attention consistency has been proved in these experiments.

\begin{table}[t]
\begin{center}
\begin{tabular}{l|cc}
\hline
\textbf{Model} & \textbf{Top-1} & \textbf{Top-5} \\
\hline
TimeSformer-L \cite{bertasius2021space} & 62.40 & 81.00 \\
SlowFast R101 \cite{feichtenhofer2019slowfast} & 63.10 &  87.60 \\
MSNet R50 \cite{kwon2020motionsqueeze} & 64.70 & 89.40 \\
bLVNet R101 \cite{fan2019more} & 65.20 & 90.30 \\
ViViT-L \cite{arnab2021vivit} & 65.90 & 89.90 \\
TSM RGB+Flow \cite{lin2019tsm} & 66.60 & 91.30 \\
MViT-B-24 \cite{fan2021multiscale} & 68.70 & 91.50 \\
Swin-B \cite{liu2022video} & 69.60 & \textbf{92.70} \\
\hline\hline
\textbf{Ours} & \textbf{70.74} & 92.18 \\
\hline
\end{tabular}
\end{center}
\caption{\textbf{Comparison to state-of-the-art methods on Something-Something V2 dataset \cite{goyal2017something}.} The proposed evaluation result is trained using directed Gromov-Wasserstein loss $\mathcal{L}_{DGW}$.}
\label{table:ssv2}
\end{table}

\section{Conclusions}

This paper presents a novel multi-view attention consistency method using directed Gromov-Wasserstein discrepancy for the action recognition explanation.
The directed Gromov-Wasserstein discrepancy not only computes the similarity between attention volumes from different views but also maintains the motion information and the structure of the compared attentions. 
Moreover, using the Neural Radiance Fields for implicit feature rendering has solved the training problem on single-view datasets.
The ablation studies on the Jester dataset have shown the effectiveness of our proposed approach. 
Particularly, the performance of the action recognition model has been notably improved by using our directed Gromov-Wasserstein loss.
Furthermore, the experimental results on the two large-scale action recognition benchmarks, i.e., Something-Something V2 and Kinetics-400, have confirmed the high accuracy performance of our proposed method.

{\small
\bibliographystyle{ieee_fullname}
\bibliography{egbib}
}

\end{document}